\documentclass[10pt,twocolumn,letterpaper]{article}

\usepackage{iccv}
\usepackage{times}
\usepackage{epsfig}
\usepackage{graphicx}
\usepackage{amsmath}
\usepackage{amssymb}
\usepackage{url}

\usepackage{upgreek}
\usepackage{booktabs}

\usepackage{pifont}         % http://ctan.org/pkg/pifont
\newcommand{\cmark}{\ding{51}}

\makeatletter
\@namedef{ver@everyshi.sty}{}
\makeatother
\usepackage{pgfplots}

\definecolor{chromeyellow}{rgb}{1.0, 0.65, 0.0}
\definecolor{darkcerulean}{rgb}{0.03, 0.27, 0.49}
\definecolor{darkorange}{rgb}{1.0, 0.55, 0.0}
\definecolor{darkmidnightblue}{rgb}{0.0, 0.2, 0.4}
\definecolor{internationalorange}{rgb}{1.0, 0.31, 0.0}
\definecolor{internationalkleinblue}{rgb}{0.0, 0.18, 0.65}
\definecolor{lightsalmon}{rgb}{1.0, 0.63, 0.48}
\definecolor{mangotango}{rgb}{1.0, 0.51, 0.26}
\definecolor{mayablue}{rgb}{0.45, 0.76, 0.98}
\definecolor{majorelleblue}{rgb}{0.38, 0.31, 0.86}
\definecolor{mediumelectricblue}{rgb}{0.01, 0.31, 0.59}
\definecolor{bananamania}{rgb}{0.98, 0.91, 0.71}
\definecolor{mossgreen}{rgb}{0.68, 0.87, 0.68}
\definecolor{mintgreen}{rgb}{0.6, 1.0, 0.6}
\definecolor{palegreen}{rgb}{0.6, 0.98, 0.6}
\definecolor{bananayellow}{rgb}{1.0, 0.88, 0.21}
\definecolor{bluebell}{rgb}{0.64, 0.64, 0.82}
\definecolor{carnationpink}{rgb}{1.0, 0.65, 0.79}

\usepackage[pagebackref=true,breaklinks=true,colorlinks,bookmarks=false]{hyperref}

\iccvfinalcopy % *** Uncomment this line for the final submission

\newlength\savewidth\newcommand\shline{\noalign{\global\savewidth\arrayrulewidth
		\global\arrayrulewidth 1pt}\hline\noalign{\global\arrayrulewidth\savewidth}}

\def\iccvPaperID{8703} % *** Enter the ICCV Paper ID here

\ificcvfinal\fi

\usepackage{pgfplots}
\usepackage{tikz}

\definecolor{hous}{HTML}{b88b4d}
\definecolor{green}{HTML}{79c561}
\definecolor{farming}{HTML}{ded94c}
\definecolor{trans}{HTML}{b4b4a9}
\definecolor{services}{HTML}{ff362e}
\definecolor{other}{HTML}{dbd4d3}
\definecolor{industry}{HTML}{db79c0}
\definecolor{water}{HTML}{7982db}
\definecolor{techinfra}{HTML}{303355}

\pgfplotsset{compat=1.17} 
\begin{document}

\title{Conditional DETR for Fast Training Convergence}
\newcommand*\samethanks[1][\value{footnote}]{\footnotemark[#1]}
\author{
Depu Meng$^{1}$\thanks{The two authors share first authorship, and the order was determined by rolling dice.
This work was done when D. Meng, X. Chen, and Z. Fan were interns at Microsoft Research, Beijing, P.R. China} \quad
Xiaokang Chen$^{2}$\samethanks \quad
Zejia Fan$^2$ \quad
Gang Zeng$^2$ \quad \\
Houqiang Li$^1$ \quad
Yuhui Yuan$^3$ \quad
Lei Sun$^3$ \quad 
Jingdong Wang$^3$\thanks{Corresponding author.} \\[1.2mm]
$^1$University of Science and Technology of China \quad
$^2$Peking University \quad
$^3$Microsoft Research Asia
}

\maketitle
\ificcvfinal\thispagestyle{empty}\fi

%%%%%%%%% ABSTRACT
\begin{abstract}
The recently-developed DETR approach
applies the transformer encoder and decoder architecture
to object detection
and
achieves promising performance.
In this paper,
we handle the critical issue, slow training convergence,
and present a conditional cross-attention mechanism
for fast DETR training.
Our approach is motivated by 
that the cross-attention in DETR relies highly on 
the content embeddings 
for localizing the four extremities
and predicting the box,
which increases the need for high-quality content embeddings
and thus the training difficulty.

Our approach,
named conditional DETR,
learns a conditional spatial query
from the decoder embedding
for decoder multi-head cross-attention.
The benefit is that
through the conditional spatial query,
each cross-attention head is able to attend
to a band containing a distinct region,
e.g., one object extremity
or a region inside the object box.
This narrows down
the spatial range for localizing the distinct regions
for object classification and box regression,
thus relaxing the dependence on 
the content embeddings and
easing the training.
Empirical results show that
conditional DETR converges $6.7\times$ faster 
for the backbones R$50$ and R$101$
and $10\times$ faster for stronger backbones 
DC$5$-R$50$ and DC$5$-R$101$.
Code is available at~\url{https://github.com/Atten4Vis/ConditionalDETR}.

\end{abstract}

\section{Introduction}
The DEtection TRansformer (DETR) method~\cite{CarionMSUKZ20}
applies the transformer encoder and decoder architecture to object detection
and achieves good performance.
It effectively eliminates the need for
many hand-crafted components,
including non-maximum suppression 
and anchor generation.

\begin{figure}[t]
\centering
\footnotesize
\includegraphics[width=.24\linewidth]{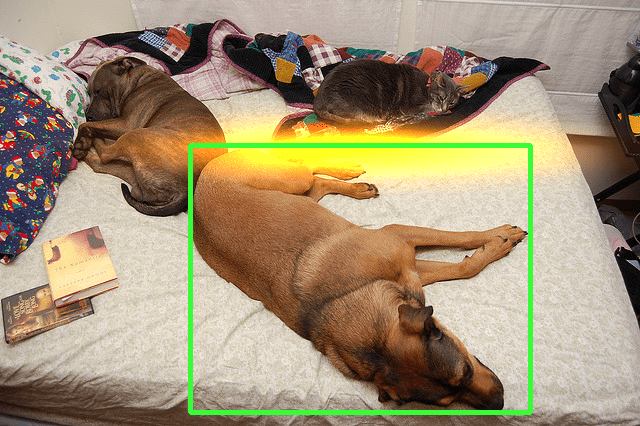}
\includegraphics[width=.24\linewidth]{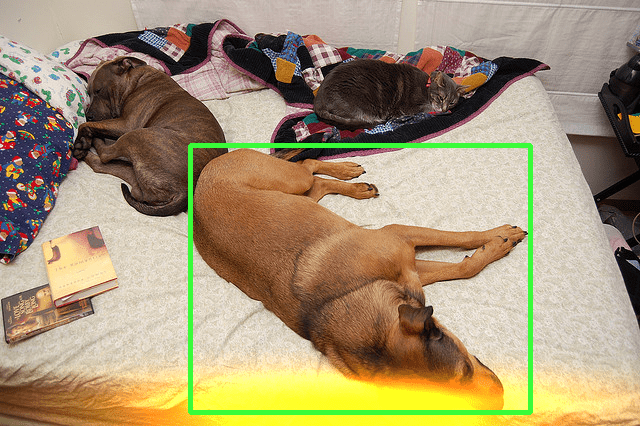}
\includegraphics[width=.24\linewidth]{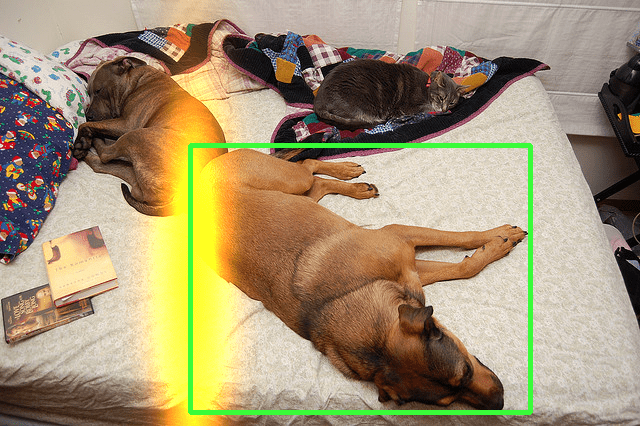}
\includegraphics[width=.24\linewidth]{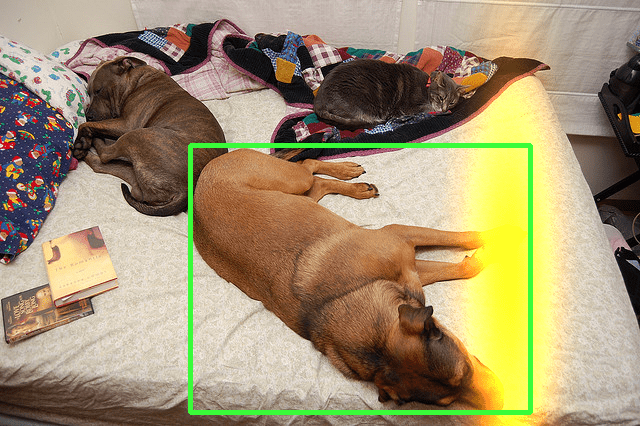}\\
\includegraphics[width=.24\linewidth]{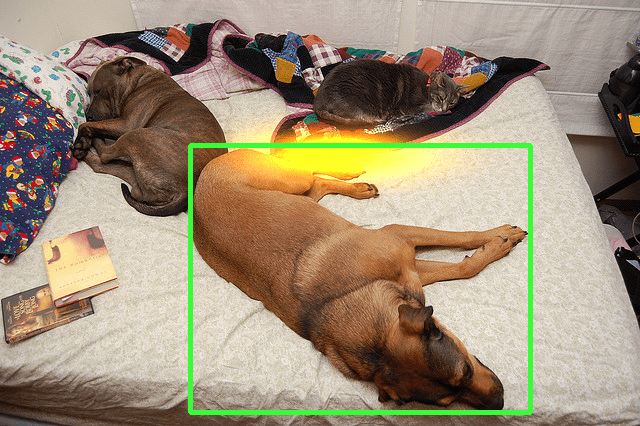}
\includegraphics[width=.24\linewidth]{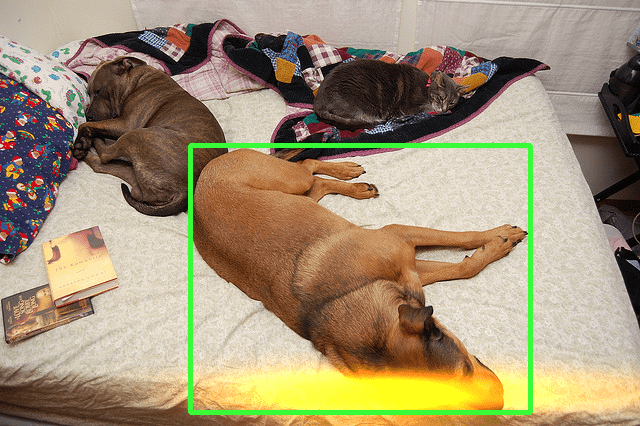}
\includegraphics[width=.24\linewidth]{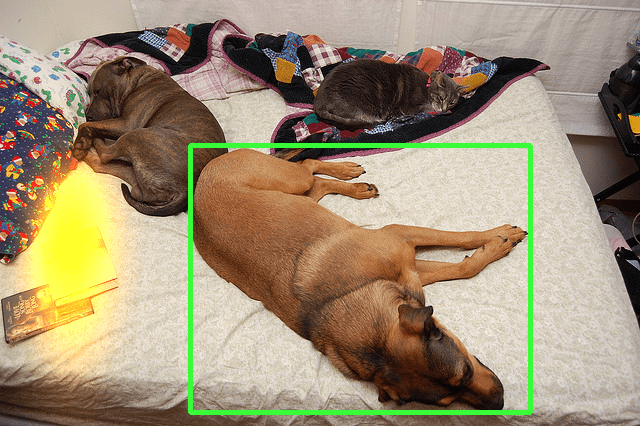}
\includegraphics[width=.24\linewidth]{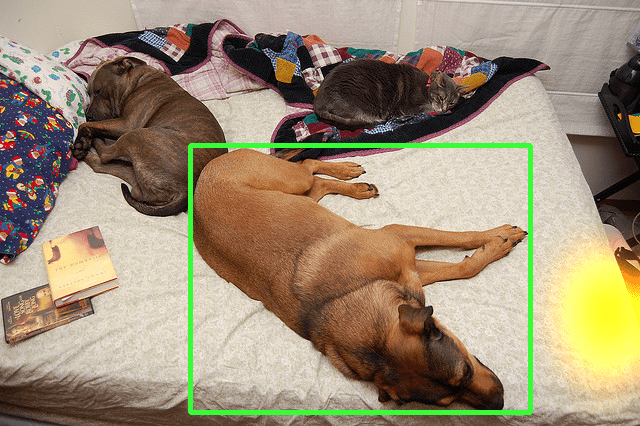}\\
\includegraphics[width=.24\linewidth]{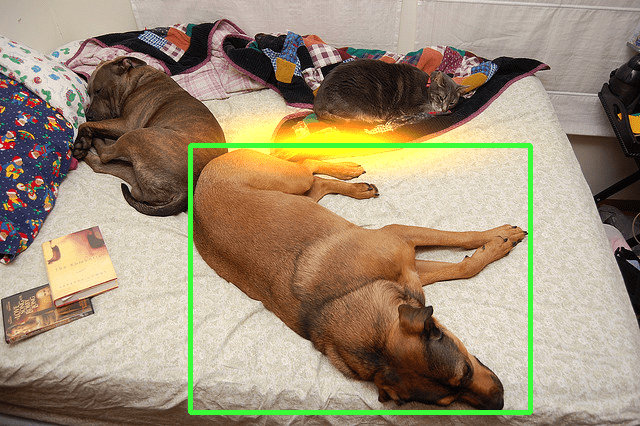}
\includegraphics[width=.24\linewidth]{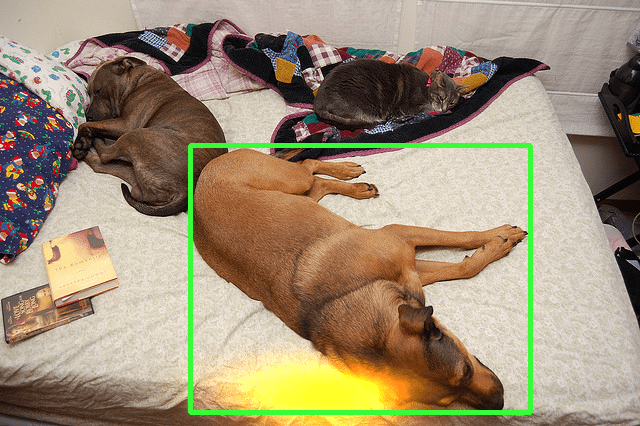}
\includegraphics[width=.24\linewidth]{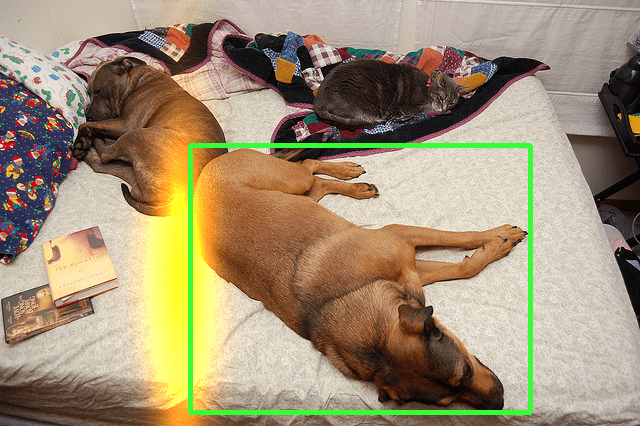}
\includegraphics[width=.24\linewidth]{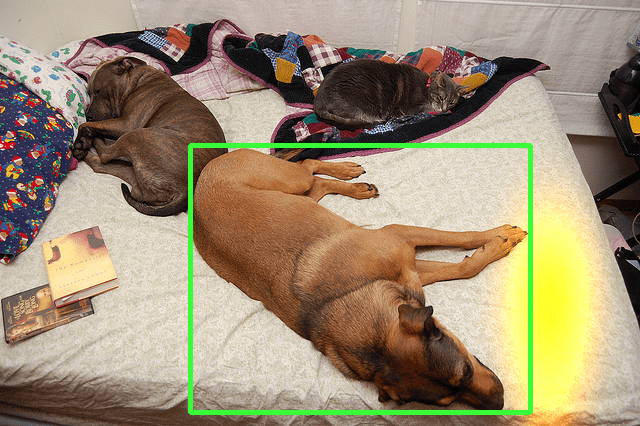}\\
   \caption{Comparison of spatial attention weight maps
   for our conditional DETR-R$50$ with $50$ training epochs (the first row),
   the original DETR-R$50$ with $50$ training epochs (the second row),
   and the original DETR-R$50$ with $500$ training epochs (the third row).
   The maps for our conditional DETR and 
   DETR trained with $500$ epochs
   are able to highlight the four extremity regions satisfactorily.
   In contrast,
   the spatial attention weight maps
   responsible for the left and right edges (the third and fourth images in the second row)
   from DETR trained with $50$ epochs
   cannot highlight the extremities satisfactorily.
   The green box is the ground-truth box.
   }
\label{fig:teaserspatialattentionmap}
\vspace{-.2cm}
\end{figure}

The DETR approach suffers from
slow convergence on training,
and needs $500$ training epochs to get good performance.
The very recent work,
deformable DETR~\cite{ZhuSLLWD20},
handles this issue
by replacing the global dense attention
(self-attention and cross-attention)
with deformable attention that 
attends to a small set of key sampling points
and using the high-resolution and multi-scale encoder.
Instead, we still use the global dense attention and 
propose an improved decoder cross-attention mechanism
for accelerating the training process.

Our approach is motivated by high dependence on content embeddings
and minor contributions made by the spatial embeddings in cross-attention.
The empirical results in DETR~\cite{CarionMSUKZ20}
show
that if removing the positional embeddings in keys and the object queries 
from the second decoder layer
and only using the content embeddings in keys and queries,
the detection AP drops slightly\footnote{
The minor AP drop $1.4$ is
reported on R$50$ with $300$ epochs in Table 3 from~\cite{CarionMSUKZ20}.
We empirically got the consistent observation:
the AP drops to $34.0$ from $34.9$
for $50$ training epochs.
}.

Figure~\ref{fig:teaserspatialattentionmap} (the second row)
shows that 
the spatial attention weight maps
from 
the cross-attention in DETR trained with $50$ epochs.
One can see that two among the four maps
do not correctly highlight the bands for the corresponding extremities,
thus weak at shrinking the spatial range
for the content queries
to precisely localize the extremities.
The reasons are that (i) the spatial queries, i.e.,
object queries,
only give the general attention weight map
without exploiting the specific image information;
and 
that (ii) due to short training 
the content queries are not strong enough 
to 
match the spatial keys well
as they are also used to match the content keys.
This
increases the dependence on high-quality content embeddings,
thus increasing the training difficulty.

\begin{figure}[t]
    \includegraphics[width=0.9\linewidth]{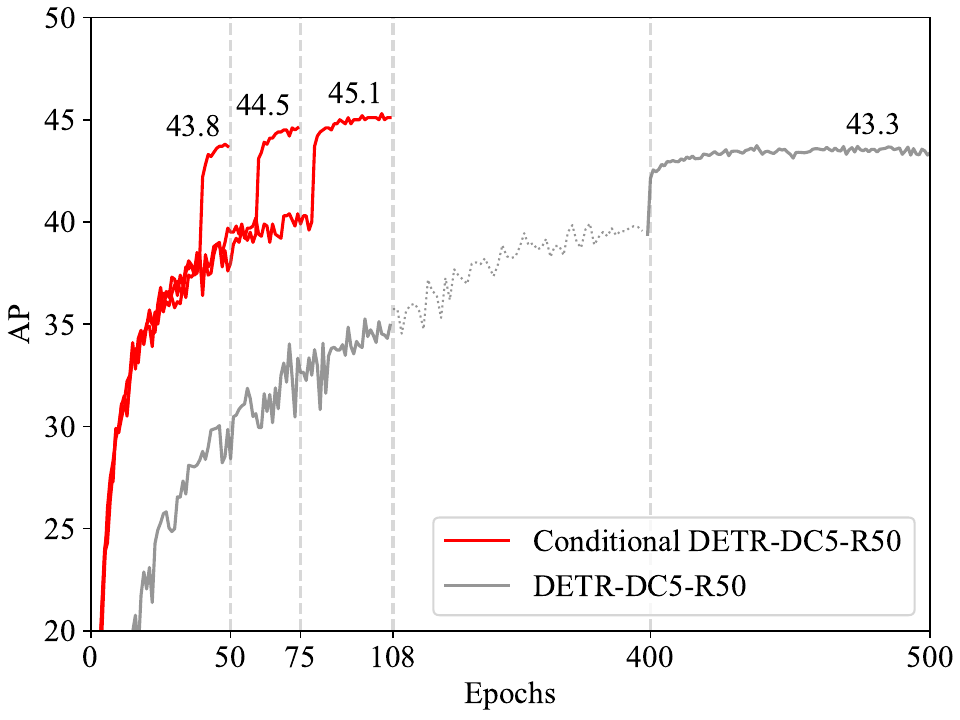}\vspace{-.15cm}
   \caption{Convergence curves
   for conditional DETR-DC5-R50
   and DETR-DC5-R50 on COCO 2017 \texttt{val}. 
   The conditional DETR is trained
   for $50$, $75$, $108$ epochs.
   Conditional DETR training is converged
   much faster than DETR.}
\label{fig:convergencecurve}
\vspace{-.3cm}
\end{figure}

We present a conditional DETR approach,
which learns a conditional spatial embedding
for each query
from the corresponding previous decoder output embedding,
to form a so-called conditional spatial query
for decoder multi-head cross-attention.
The conditional spatial query
is predicted 
by 
mapping
the information for regressing the object box
to the embedding space,
the same to the space
that the $2$D coordinates of the keys
are also mapped to.

We empirically observe that
using the spatial queries and keys,
each cross-attention head
spatially attends to 
a band
containing the object extremity
or a region inside the object box
(Figure~\ref{fig:teaserspatialattentionmap},
the first row).
This shrinks
the spatial range for the content queries
to localize the effective regions
for class and box prediction.
As a result, 
the dependence on the content embeddings is relaxed
and the training is easier.
The experiments show that
conditional DETR converges $6.7\times$ faster 
for the backbones R$50$ and R$101$
and $10\times$ faster for stronger backbones
DC$5$-R$50$ and DC$5$-R$101$. 
Figure~\ref{fig:convergencecurve}
gives the convergence curves for
conditional DETR and
the original DETR~\cite{CarionMSUKZ20}.

\section{Related Work}
\noindent\textbf{Anchor-based and anchor-free detection.}
Most existing object detection approaches
make predictions from initial guesses
that are carefully designed. 
There are two main
initial guesses: 
anchor boxes
or object centers.
The anchor box-based methods inherit the ideas
from the proposal-based method, Fast R-CNN.
Example methods include
Faster R-CNN~\cite{Girshick15},
SSD~\cite{LiuAESRFB16},
YOLOv2~\cite{RedmonF17},
YOLOv3~\cite{JA18},
YOLOv4~\cite{BWL20},
RetinaNet~\cite{LinGGHD20},
Cascade R-CNN~\cite{CaiV18},
Libra R-CNN~\cite{PangCSFOL19},
TSD~\cite{SongLW20}
and so on.

The anchor-free detectors predict the boxes
at points near the object centers.
Typical methods include
YOLOv1~\cite{RedmonDGF16},
CornerNet~\cite{LawD18},
ExtremeNet~\cite{ZhouZK19},
CenterNet~\cite{ZWP19,DBXQHT19},
FCOS~\cite{TianSCH19}
and others~\cite{LiCWZ19,LuLYLY19,ZhuHS19,KongSLJS19,ZhuCSS20,LawTRD20,HuangYDY15,YuJWCH16,zhang2020bridging}.

\vspace{.1cm}
\noindent\textbf{DETR and its variants.}
DETR successfully applies transformers
to object detection,
effectively removing the need for many hand-designed components
like non-maximum suppression
or initial guess generation.
The high computation complexity issue,
caused by the global encoder self-attention,
is handled in adaptive clustering transformer~\cite{ZhengGWLD20} 
and by sparse attentions in deformable DETR~\cite{ZhuSLLWD20}.

The other critical issue, slow training convergence,
has been attracting a lot of recent research attention.
The TSP (transformer-based set prediction) 
approach~\cite{SunCYK20} 
eliminates the cross-attention modules
and combines the FCOS and R-CNN-like detection heads.
Deformable DETR~\cite{ZhuSLLWD20}
adopts deformable attention,
which attends to sparse positions learned from the content embedding,
to replace decoder cross-attention.

The spatially modulated co-attention (SMCA) approach~\cite{GaoZWDL21},
which is~{concurrent} to our approach,
is very close to our approach.
It
modulates the DETR multi-head global cross-attentions
with Gaussian maps
around a few (shifted) centers that are learned from the decoder embeddings,
to focus more on a few regions
inside the estimated box.
In contrast,
the proposed conditional DETR approach
learns the conditional spatial queries from the decoder content embeddings,
and predicts the spatial attention weight maps
without human-crafting the attention attenuation,
which highlight four extremities
for box regression,
and distinct regions inside the object for classification.

\vspace{.1cm}
\noindent\textbf{Conditional and dynamic convolution.}
The proposed conditional spatial query scheme
is related to conditional convolutional kernel generation. 
Dynamic filter network~\cite{JiaBTG16}
learns the convolutional kernels from the input,
which is applied to
instance segmentation
in CondInst~\cite{TianSC20} and SOLOv2~\cite{WangZKLS20}
for learning instance-dependent convolutional kernels.
CondConv~\cite{YangBLN19} and
dynamic convolution~\cite{ChenDLCYL20}
mix convolutional kernels
with the weights 
learned from the input.
SENet~\cite{HuSS18}, 
GENet~\cite{HuSASV18}
abd Lite-HRNet~\cite{YuXGYZSW21}
learn from the input
the channel-wise weights.

These methods learn from the input
the convolutional kernel weights 
and then apply the convolutions to the input.
In contrast,
the linear projection in our approach is learned
from the decoder embeddings
for representing the displacement and scaling information.

\vspace{.1cm}
\noindent\textbf{Transformers.}
The transformer~\cite{VaswaniSPUJGKP17} relies on
the attention mechanism, self-attention and cross-attention,
to draw global dependencies 
between the input and the output.
There are several works closely related to our approach.
Gaussian transformer~\cite{GuoZL19}
and T-GSA (Transformer with Gaussian-weighted self-attention)~\cite{KimEL20},
followed by SMCA~\cite{GaoZWDL21},
attenuate the attention weights according to
the distance between target and context symbols
with learned or human-crafted Gaussian variance.
Similar to ours,
TUPE~\cite{KeHL20} computes the attention weight
also from the spatial attention weight and the content attention weight.
Instead,
our approach mainly focuses on the attention attenuation mechanism
in a learnable form other than a Gaussian function,
and potentially benefits speech enhancement~\cite{KimEL20}
and natural language inference~\cite{GuoZL19}.

\section{Conditional DETR}
\subsection{Overview}
\noindent\textbf{Pipeline.}
The proposed approach
follows detection transformer (DETR), an end-to-end object detector,
and predicts all the objects at once
without the need for NMS or anchor generation.
The architecture consists of
a CNN backbone,
a transformer encoder,
a transformer decoder,
and object class and box position predictors.
The transformer encoder aims to improve the content embeddings output from the CNN backbone.
It is a stack
of multiple encoder layers,
where each layer mainly consists of 
a self-attention layer
and a feed-forward layer.

\begin{figure}[t]
\centering
\includegraphics[width=.4\textwidth]{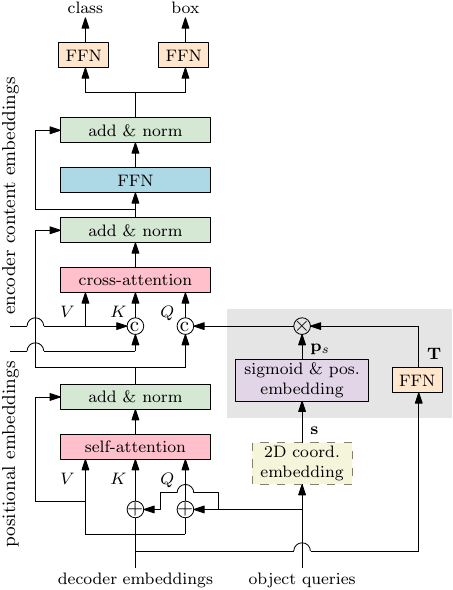}
 
   \caption{Illustrating
   one decoder layer in conditional DETR.
   The main difference from the original DETR~\cite{CarionMSUKZ20} 
   lies in the input queries and the input keys 
   for cross-attention.
   The conditional spatial query
   is predicted
   from learnable $2$D coordinates $\mathbf{s}$
   and the embeddings
   output from the previous decoder layer,
   through the operations depicted in the gray-shaded box.
   The $2$D coordinate $\mathbf{s}$
   can be predicted from the object query (the dashed box),
   or simply learned as model parameters
   The spatial query (key) and the content query (key)
   are concatenated as the query (key).
   The resulting cross-attention is called
   conditional cross-attention.
   Same as DETR~\cite{CarionMSUKZ20},
   the decoder layer is repeated $6$ times.
   }
\label{fig:ConditionalDETRArchitecture}
\vspace{-.2cm}
\end{figure}

The transformer decoder
is a stack of decoder layers.
Each decoder layer,
illustrated in Figure~\ref{fig:ConditionalDETRArchitecture},
is composed of three main layers:
(1) a self-attention layer
for removing duplication prediction,
which 
performs interactions between the embeddings,
outputted from the previous decoder layer and used for class and box prediction,
(2) a cross-attention layer,
which aggregates the embeddings output from the encoder
to refine the decoder embeddings for improving class and box prediction,
and (3) a feed-forward layer.

\vspace{.1cm}
\noindent\textbf{Box regression.}
A candidate box 
is predicted
from each decoder embedding as follows,
\begin{align}
    \mathbf{b} = \operatorname{sigmoid}(\operatorname{FFN}(\mathbf{f}) + [{\mathbf{s}}^\top~0~0]^\top)
    \label{eqn:boxprediction}.
\end{align}
Here, $\mathbf{f}$ is the decoder embedding.
$\mathbf{b}$ is a four-dimensional vector
$[b_{cx}~b_{cy}~b_{w}~b_{h}]^\top$,
consisting of
the box center,
the box width
and the box height.
$\operatorname{sigmoid}()$ is used to normalize
the prediction $\mathbf{b}$
to the range $[0, 1]$.
$\operatorname{FFN}()$
aims to predictthe unnormalized box.
${\mathbf{s}}$
is the unnormalized $2$D coordinate of 
the reference point,
and is $(0,0)$
in the original DETR.
In our approach,
we consider two choices:
learn the reference point $\mathbf{s}$ as a parameter for each candidate box prediction,
or generate it from the corresponding object query.

\vspace{.1cm}
\noindent\textbf{Category prediction.}
The classification score for each candidate box
is also predicted from
the decoder embedding
through an FNN,
$\mathbf{e} = \operatorname{FFN}(\mathbf{f})
$.

\vspace{.1cm}
\noindent\textbf{Main work.}
The cross-attention mechanism
aims to \emph{localize the distinct regions,
four extremities for box detection 
and regions inside the box for object classification,
and aggregates the corresponding embeddings}.
We propose a conditional cross-attention mechanism 
with introducing conditional spatial queries
for
improving the localization capability
and accelerating the training process.

\subsection{DETR Decoder Cross-Attention}
The DETR decoder cross-attention mechanism
takes three inputs:
queries,
keys and values. 
Each key is formed 
by adding
a content key $\mathbf{c}_k$ 
(the content embedding output from the encoder) and 
a spatial key $\mathbf{p}_k$
(the positional embedding of the corresponding normalized $2$D coordinate).
The value is formed from the content embedding, same with the content key, output from the encoder.

In the original DETR approach,
each query 
is formed 
by adding
a content query $\mathbf{c}_q$
(the embedding output from the decoder self-attention),
and 
a spatial query $\mathbf{p}_q$
(i.e., the object query $\mathbf{o}_q$).
In our implementation, there are $N=300$ object queries,
and accordingly there are $N$ queries\footnote{For description simplicity and clearness,
we drop the query, key, and value indices.},
each query outputting a candidate detect result in one decoder layer.

The attention weight is based on
the dot-product between the query and the key, used for attention weight computation, \begin{align}
    &(\mathbf{c}_q + \mathbf{p}_q)^\top (\mathbf{c}_k + \mathbf{p}_k) \nonumber\\
    =~&\mathbf{c}_q^\top \mathbf{c}_k 
    + \mathbf{c}_q^\top \mathbf{p}_k
    + \mathbf{p}_q^\top \mathbf{c}_k
    + \mathbf{p}_q^\top \mathbf{p}_k \nonumber \\
    =~&\mathbf{c}_q^\top \mathbf{c}_k 
    + \mathbf{c}_q^\top \mathbf{p}_k
    + \mathbf{o}_q^\top \mathbf{c}_k
    + \mathbf{o}_q^\top \mathbf{p}_k.
\end{align}

\subsection{Conditional Cross-Attention}
\label{sec:conditionalcrossattention}
The proposed conditional cross-attention mechanism
forms the query 
by concatenating 
the content query $\mathbf{c}_q$,
outputting from decoder self-attention,
and the spatial query $\mathbf{p}_q$.
Accordingly,
the key is formed as the concatenation 
of the content key $\mathbf{c}_k$
and the spatial key $\mathbf{p}_k$.

The cross-attention weights
consist of two components,
content attention weight
and spatial attention weight.
The two weights
are from two dot-products,
content and spatial
dot-products,
\begin{align}
\mathbf{c}_q^\top\mathbf{c}_k
+ \mathbf{p}_q^\top\mathbf{p}_k.
\end{align} 
Different from the original DETR cross-attention,
our mechanism separates the roles of content and spatial queries
so that spatial and content queries focus on the spatial and content attention weights, respectively.

An additional important task 
is to compute
the spatial query $\mathbf{p}_q$
from the embedding $\mathbf{f}$
of the previous decoder layer.
We first identify
that
the spatial information
of the distinct regions
are determined by
the two factors together,
decoder embedding
and reference point.
We then show how to map them
to the embedding space,
forming the query $\mathbf{p}_q$,
so that 
the spatial query lies in the same space
the $2$D coordinates of the keys
are mapped to.

\vspace{.1cm}
\noindent\textbf{The decoder embedding
contains the displacements}
of the distinct regions with respect to 
the reference point.
The box prediction process in Equation~\ref{eqn:boxprediction}
consists of two steps:
(1) predicting the box 
with respect to the reference point
in the unnormalized space,
and (2) normalizing the predicted box 
to the range $[0, 1]$\footnote{The origin $(0,0)$
in the unnormalized space 
for the original DETR method
is mapped to $(0.5, 0.5)$
(the center in the image space)
in the normalized space 
through the $\operatorname{sigmoid}$ function.}.

Step (1) means that
the decoder embedding $\mathbf{f}$
contains the displacements
of the four extremities
(forming the box)
with respect to the reference point $\mathbf{s}$
in the unnormalized space.
This implies that
both the embedding $\mathbf{f}$
and the reference point $\mathbf{s}$
are necessary to determine
the spatial information
of the distinct regions,
the four extremities
as well as the region for predicting the classification score.

\vspace{.1cm}
\noindent\textbf{Conditional spatial query prediction.}
We predict the conditional spatial query
from the embedding $\mathbf{f}$
and the reference point $\mathbf{s}$,
\begin{align}
    (\mathbf{s}, \mathbf{f}) \rightarrow \mathbf{p}_q,
\end{align}
so that it is aligned with the positional space which the normalized $2$D coordinates
of the keys are mapped to.
The process
is 
illustrated in
the gray-shaded box area of Figure~\ref{fig:ConditionalDETRArchitecture}.

We normalize the reference point $\mathbf{s}$
and then map it to a $256$-dimensional sinusoidal positional embedding
in the same way as the positional embedding
for keys:
\begin{align}
    \mathbf{p}_s = 
    \operatorname{sinusoidal}(\operatorname{sigmoid}(\mathbf{s})).
\end{align}
We then map the displacement information contained
in the decoder embedding $\mathbf{f}$
to a linear projection in the same space
through an FFN
consisting of learnable linear projection + ReLU + learnable linear projection:
$\mathbf{T} = \operatorname{FFN}(\mathbf{f})$.

The conditional spatial query is computed
by transforming the reference point
in the embedding space:
$\mathbf{p}_q = \mathbf{T} \mathbf{p}_s$.
We choose the simple and computationally-efficient projection matrix,
a diagonal matrix.
The $256$ diagonal elements 
are denoted as a vector $\boldsymbol{\uplambda}_q$.
The conditional spatial query is computed
by the element-wise multiplication:
\begin{align}
    \mathbf{p}_q =  \mathbf{T} {\mathbf{p}}_s = \boldsymbol{\uplambda}_q \odot {\mathbf{p}}_s.\label{eqn:conditionalspatialembedding}
\end{align}

\vspace{.1cm}
\noindent\textbf{Multi-head cross-attention.}
Following
DETR~\cite{CarionMSUKZ20},
we adopt the standard multi-head cross-attention mechanism.
Object detection usually needs
to implicitly or explicitly localize the four object extremities for
accurate box regression
and localize the object region for 
accurate object classification.
The multi-head mechanism is beneficial 
to disentangle the localization tasks.

We perform multi-head parallel attentions
by projecting the queries,
the keys,
and the values
$M=8$ times
with learned linear projections
to low dimensions.
The spatial and content queries (keys)
are separately projected to each head with different linear projections.
The projections for values are the same as the original DETR
and are only for the contents.

\begin{figure*}[t]
\centering
\includegraphics[ height=.087\linewidth]{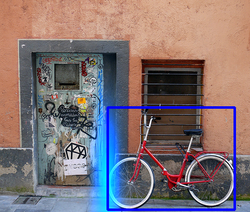}
\includegraphics[ height=.087\linewidth]{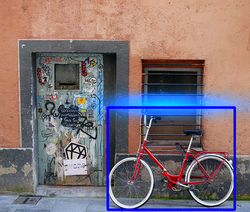}
\includegraphics[ height=.087\linewidth]{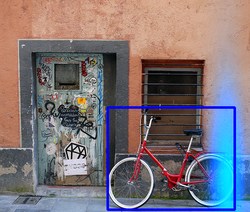}
\includegraphics[ height=.087\linewidth]{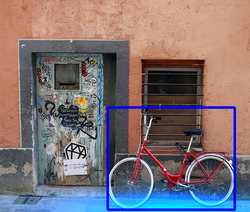}
\includegraphics[ height=.087\linewidth]{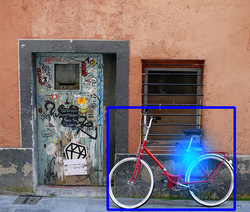}~
\includegraphics[ height=.087\linewidth, trim={1cm 0 2.75cm 0},clip]{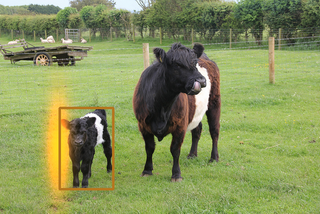}
\includegraphics[ height=.087\linewidth, trim={1cm 0 2.75cm 0},clip]{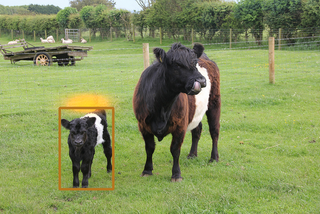}
\includegraphics[ height=.087\linewidth, trim={1cm 0 2.75cm 0},clip]{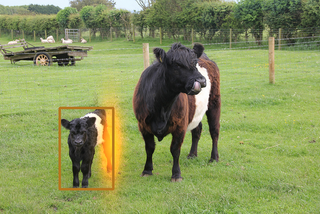}
\includegraphics[ height=.087\linewidth, trim={1cm 0 2.75cm 0},clip]{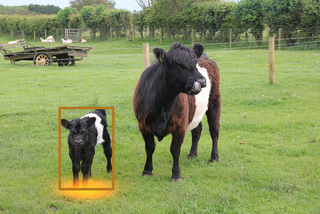}
\includegraphics[ height=.087\linewidth, trim={1cm 0 2.75cm 0},clip]{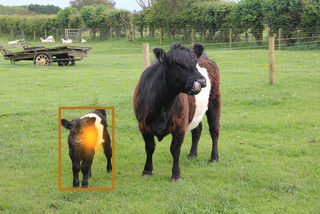}
\\\vspace{.04cm}
\includegraphics[ height=.087\linewidth]{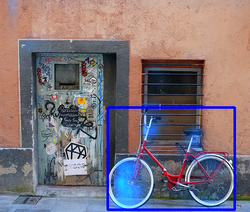}
\includegraphics[ height=.087\linewidth]{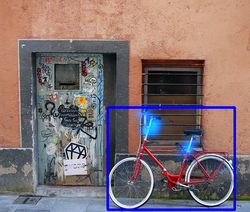}
\includegraphics[ height=.087\linewidth]{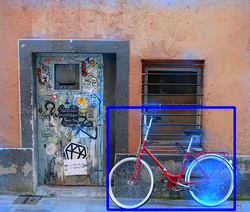}
\includegraphics[ height=.087\linewidth]{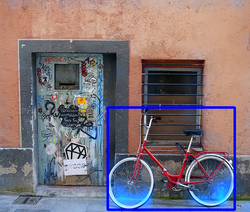}
\includegraphics[ height=.087\linewidth]{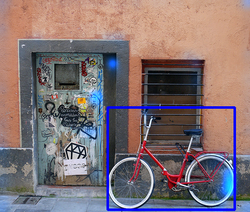}~
\includegraphics[ height=.087\linewidth, trim={1cm 0 2.75cm 0},clip]{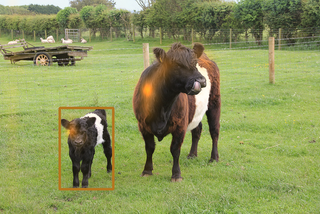}
\includegraphics[ height=.087\linewidth, trim={1cm 0 2.75cm 0},clip]{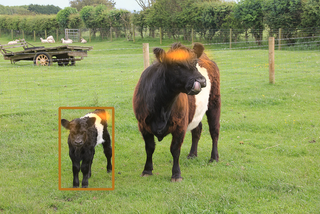}
\includegraphics[ height=.087\linewidth, trim={1cm 0 2.75cm 0},clip]{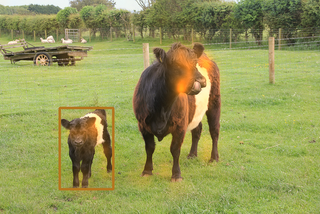}
\includegraphics[ height=.087\linewidth, trim={1cm 0 2.75cm 0},clip]{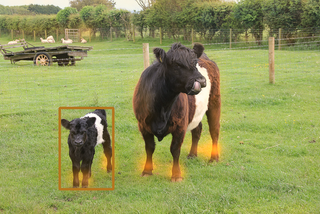}
\includegraphics[ height=.087\linewidth, trim={1cm 0 2.75cm 0},clip]{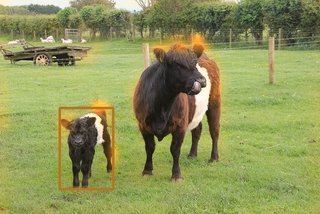}
\\\vspace{.04cm}
\includegraphics[ height=.087\linewidth]{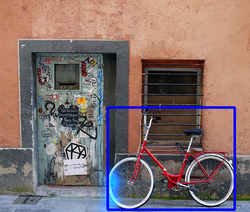}
\includegraphics[ height=.087\linewidth]{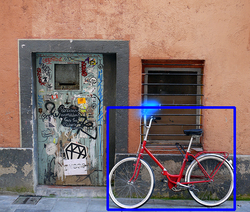}
\includegraphics[ height=.087\linewidth]{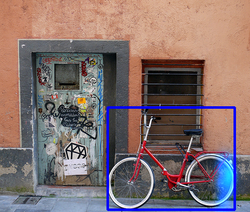}
\includegraphics[ height=.087\linewidth]{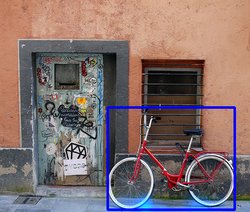}
\includegraphics[ height=.087\linewidth]{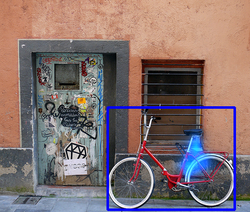}~
\includegraphics[ height=.087\linewidth, trim={1cm 0 2.75cm 0},clip]{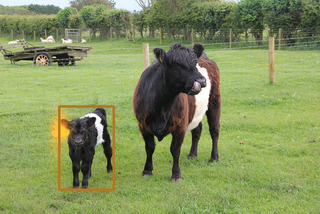}
\includegraphics[ height=.087\linewidth, trim={1cm 0 2.75cm 0},clip]{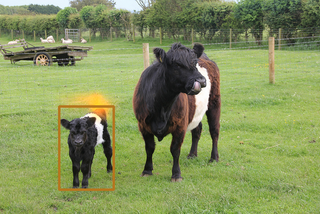}
\includegraphics[ height=.087\linewidth, trim={1cm 0 2.75cm 0},clip]{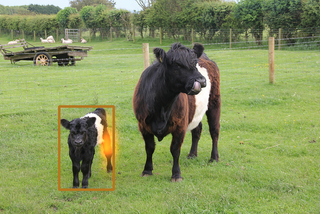}
\includegraphics[ height=.087\linewidth, trim={1cm 0 2.75cm 0},clip]{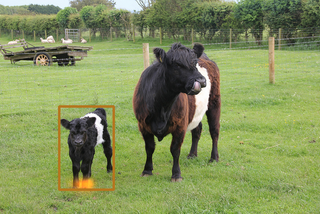}
\includegraphics[ height=.087\linewidth, trim={1cm 0 2.75cm 0},clip]{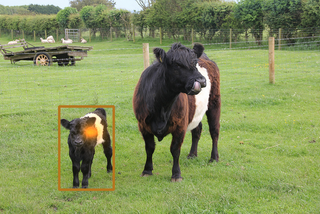}
\\
   \caption{
   Illustrating the spatial attention weight maps (the first row),
   the content attention weight maps (the second row),
   and the combined attention weight maps (the third row)
   computed from our conditional DETR.
   The attention weight maps 
   are from $5$ heads out of the $8$ heads
   and are responsible for
   the four extremities and a region inside the object box. 
   The content attention weight maps
   for the four extremities highlight
   scattered regions inside the box (bicycle)
   or similar regions in two object instances 
   (cow),
   and 
   the corresponding combined attention weight maps
   highlight the extremity regions
   with the help of the spatial attention weight maps.
   The combined attention weight map
   for the region inside the object box
   mainly depends on the spatial attention weight map,
   which implies that the representation
   of a region inside the object might 
   encode enough class information.
   The maps are from conditional DETR-R$50$ trained with $50$ epochs.
   }
\label{fig:multiheadcombinedattention}
\vspace{-.3cm}
\end{figure*}

\subsection{Visualization and Analysis}
\noindent\textbf{Visualization.}
Figure~\ref{fig:multiheadcombinedattention}
visualizes the attention weight maps
for each head:
the spatial attention weight maps,
the content attention weight maps,
and the combined attention weight maps.
The maps are soft-max normalized 
over the spatial dot-products $\mathbf{p}_q^\top\mathbf{p}_k$,
the content dot-products $\mathbf{c}_q^\top\mathbf{c}_k$,
and 
the combined dot-products
$\mathbf{c}_q^\top\mathbf{c}_k
+ \mathbf{p}_q^\top\mathbf{p}_k$.
We show $5$ out of the $8$ maps, 
and other three are the duplicates,
corresponding to bottom and top extremities, 
and a small region inside the object box\footnote{The duplicates might be different for models trained several times,
but the detection performance is almost the same.}. 

We can see that
the spatial attention weight map at each head is able to 
localize a distinct region,
a region containing one extremity or 
a region inside the object box.
It is interesting that
each spatial attention weight map corresponding to an extremity
highlights a spatial band 
that overlaps with the corresponding edge of the object box.
The other spatial attention map for the region inside the object box
merely highlights a small region
whose representations might already encode enough information for object classification.

The content attention weight maps 
of the four heads corresponding to the four extremities 
highlight scattered regions in addition to the extremities.
The combination of the spatial and content maps
filters out other highlights
and keeps extremity highlights
for accurate box regression.

\vspace{.1cm}
\noindent\textbf{Comparison to DETR.}
Figure~\ref{fig:teaserspatialattentionmap}
shows the spatial attention weight maps 
of our conditional DETR (the first row)
and the original DETR trained with $50$ epochs
(the second row).
The maps of our approach are computed 
by soft-max normalizing the dot-products
between spatial keys and queries,
$\mathbf{p}_q^\top \mathbf{p_k}$.
The maps for DETR are
computed by soft-max normalizing
the dot-products with the spatial keys, 
$(\mathbf{o}_q + \mathbf{c}_q)^\top\mathbf{p}_k$.

It can be seen that
our spatial attention weight maps accurately localize the distinct regions,
four extremities.
In contrast, the maps from the original DETR with $50$ epochs can not accurately localize two extremities,
and $500$ training epochs (the third row) make the content queries stronger,
leading to accurate localization.
This implies that
it is really hard to learn the content query $\mathbf{c}_q$
to serve as two roles\footnote{Strictly speaking,
the embedding output from decoder self-attention
for more training epochs
contains both spatial and content information.
For discussion convenience, 
we still call it content query.}:
match the content key and the spatial key
simultaneously,
and thus more training epochs are needed.

\vspace{.1cm}
\noindent\textbf{Analysis.}
The spatial attention weight maps shown in Figure~\ref{fig:multiheadcombinedattention}
imply that 
the conditional spatial query,
used to form the spatial query,
have at least two effects.
(i) Translate the highlight positions to the four extremities
and the position inside the object box:
interestingly the highlighted positions
are spatially similarly distributed
in the object box.
(ii) 
Scale the spatial spread for the extremity highlights:
large spread for large objects
and small spread for small objects.

The two effects are realized
in the spatial embedding space
through applying the transformation $\mathbf{T}$
over $\mathbf{p}_s$
(further disentangled through image-independent linear projections
contained in cross-attention and distributed to each head).
This indicates that 
the transformation $\mathbf{T}$
not only contains the displacements as discussed before,
but also the object scale.

\newcommand{\blue}[1]{\textcolor{blue}{#1}}
\begin{table*}[t]
    \centering\setlength{\tabcolsep}{9.4pt}
        \footnotesize
            \renewcommand{\arraystretch}{1.3}
    \caption{Comparison of conditional DETR with DETR on COCO 2017 \texttt{val}.
    Our conditional DETR approach for high-resolution backbones DC$5$-R$50$
    and DC$5$-R$101$ is $10\times$ faster than the original DETR,
    and for low-resolution backbones 
    R$50$ and R$101$
    $6.67\times$ faster.
    Conditional DETR is empirically superior to 
    other two single-scale DETR variants.
    $^*$The results of deformable DETR are 
    from the GitHub repository
    provided by the authors of deformable DETR~\cite{ZhuSLLWD20}.
    }
    \label{tab:comparisontodetr}
    \begin{tabular}{lc|cccccccc}
        \shline
        Model & \#epochs & GFLOPs & \#params (M) & AP & AP$_{50}$ & AP$_{75}$ & AP$_{S}$ & AP$_{M}$ & AP$_{L}$\\
        \shline
        {DETR}-R$50$ & $500$ & $86$ & $41$ & $42.0$  & $62.4$  & $44.2$  & $20.5$  & $45.8$  & $61.1$ \\
        {DETR}-R$50$ & $50$ & $86$ & $41$ & $34.9$ &  $55.5$ &  $36.0$ &  $14.4$ &  $37.2$ &  $54.5$ \\
        Conditional DETR-R50 & $50$ & $90$ & $44$ & $40.9$ & $61.8$ & $43.3$ & $20.8$ & $44.6$ & $59.2$ \\
        Conditional DETR-R50 & $75$ & $90$ & $44$ & $42.1$ & $62.9$ & $44.8$ & $21.6$ & $45.4$ & $60.2$ \\
        Conditional DETR-R50 & $108$ & $90$ & $44$ & $43.0$ & $64.0$ &  $45.7$ &  $22.7$ &  $46.7$ &  $61.5$ \\
        
        \hline
        {DETR}-DC5-R$50$ & $500$ & $187$ & $41$ & $43.3$  & $63.1$  & $45.9$  & $22.5$  & $47.3$  & $61.1$ \\
        {DETR}-DC5-R$50$ & $50$ & $187$ & $41$ & $36.7$ & $57.6$ & $38.2$ & $15.4$ & $39.8$ & $56.3$ \\
        Conditional DETR-DC5-R50 & $50$ & $195$ & $44$ &  $43.8$ & $64.4$ &  $46.7$ &  $24.0$ & $47.6$ &  $60.7$  \\
        Conditional DETR-DC5-R50  & $75$ & $195$ & $44$ &  $44.5$ & $65.2$ &  $47.3$ &  $24.4$ & $48.1$ &  $62.1$  \\
        Conditional DETR-DC5-R50   & $108$ & $195$ & $44$ & $45.1$   & $65.4$   & $48.5$   & $25.3$   & $49.0$   & $62.2$   \\

        \hline
        {DETR}-R$101$ & $500$ & $152$ & $60$ & $43.5$  & $63.8$  & $46.4$  & $21.9$  & $48.0$  & $61.8$ \\
        {DETR}-R$101$ & $50$ & $152$ & $60$ & $36.9$ & $57.8$ & $38.6$ & $15.5$ & $40.6$ & $55.6$ \\
        Conditional DETR-R101  & $50$ & $156$ & $63$  & $42.8$  & $63.7$  & $46.0$  & $21.7$ & $46.6$ & $60.9$ \\
        Conditional DETR-R101   & $75$ & $156$ & $63$ & $43.7$   & $64.9$   &  $46.8$   &  $23.3$   &  $48.0$   &  $61.7$   \\
        Conditional DETR-R101   & $108$ & $156$ & $63$  & $44.5$   & $65.6$   & $47.5$   & $23.6$   & $48.4$   & $63.6$   \\
        
        \hline
        DETR-DC5-R$101$  & $ 500 $ & $ 253 $ & $ 60 $ & $ {44.9}  $ & $ {64.7}  $ & $ 47.7  $ & $ 23.7  $ & $ {49.5}  $ & $ {62.3} $\\
        DETR-DC5-R$101$  & $ $50$ $ & $ 253 $ & $ 60 $ & $ 38.6 $ & $ 59.7 $ & $ 40.7 $ & $ 17.2 $ & $ 42.2 $ & $ 57.4 $\\
        Conditional DETR-DC5-R101 & $50$ & $262$ & $63$ & $45.0$ &  $65.5$ & $48.4$ & $26.1$ & $48.9$ & $62.8$  \\
        Conditional DETR-DC5-R101    & $ $75$ $ & $ 262 $ & $ 63 $ & $ 45.6   $ & $ 66.5   $ & $ 48.8   $ & $ 25.5   $ & $ 49.7   $ & $ 63.3  $ \\
        Conditional DETR-DC5-R101   & $ $108$ $ & $ 262 $ & $ 63 $ & $ 45.9   $ & $ 66.8   $ & $ 49.5   $ & $ 27.2    $ & $ 50.3   $ & $ 63.3  $\\
        \hline
        \emph{Other single-scale DETR variants} \\
        \hline
         Deformable DETR-R50-SS$^*$  & $50$ & $ 78 $ & $ 34 $ & $ 39.4 $ & $ 59.6 $ & $ 42.3 $ & $ 20.6 $ & $ 43.0 $ & $ 55.5 $ \\
        UP-DETR-R$50$~\cite{DaiCLC20} & $ 150 $ & $ 86 $ & $ 41 $ & $ 40.5 $ & $ 60.8 $ & $ 42.6 $ & $ 19.0 $ & $ 44.4 $ & $ 60.0 $ \\
        UP-DETR-R$50$~\cite{DaiCLC20} & $ 300 $ & $ 86 $ & $ 41 $ & $ 42.8  $ & $ 63.0 $ & $ 45.3 $ & $ 20.8 $ & $ 47.1 $ & $ 61.7 $\\
        \hline
        Deformable DETR-DC5-R50-SS$^*$  & $ $50$ $ & $ 128 $ & $ 34 $ & $ 41.5 $ & $ 61.8 $ & $ 44.9 $ & $ 24.1 $ & $ 45.3 $ & $ 56.0 $ \\
        \shline
    \end{tabular}
    \vspace{-.3cm}
\end{table*}

\vspace{.1cm}
\subsection{Implementation Details}
\noindent\textbf{Architecture.}
Our architecture is almost the same with
the DETR architecture~\cite{CarionMSUKZ20}
and contains the CNN backbone,
transformer encoder,
transformer decoder,
prediction feed-forward networks (FFNs)
following each decoder layer
(the last decoder layer
and the $5$ internal decoder layers)
with parameters shared among the $6$ prediction FFNs.
The hyper-parameters are the same as DETR.

The main architecture difference is
that we introduce 
the conditional spatial embeddings 
as the spatial queries
for
conditional multi-head cross-attention
and 
that
the spatial query (key) and the content query (key)
are combined through
concatenation other than
addition.
In the first cross-attention layer
there are no decoder content embeddings,
we make simple changes based on the DETR implementation~\cite{CarionMSUKZ20}:
concatenate the positional embedding predicted from the object query
(the positional embedding)
into the original query (key).

\vspace{.1cm}
\noindent\textbf{Reference points.}
In the original DETR approach, 
${\mathbf{s}}
= [0~0]^\top$
is the same for all the decoder embeddings.
We study two ways forming 
the reference points:
regard the unnormalized $2$D coordinates
as learnable parameters,
and 
the unnormalized $2$D coordinate predicted 
from the object query
$\mathbf{o}_q$.
In the latter way
that is similar to 
deformable DETR~\cite{ZhuSLLWD20},
the prediction unit is 
an FFN
and consists of learnable linear projection + ReLU + learnable linear projection:
$\mathbf{s} = \operatorname{FFN}(\mathbf{o}_q)$.
When used for forming the conditional spatial query,
the $2$D coordinates are normalized by
the sigmoid function.

\vspace{.1cm}
\noindent\textbf{Loss function.}
We follow DETR~\cite{CarionMSUKZ20}
to find an optimal bipartite matching~\cite{Kuhn10}
between the predicted and ground-truth objects
using the Hungarian algorithm,
and then form the loss function for 
computing and back-propagate the gradients.
We use the same way with deformable DETR~\cite{ZhuSLLWD20}
to formulate the loss:
the same matching cost function, 
the same loss function
with $300$ object queries,
and the same trade-off parameters;
The classification loss function is focal loss~\cite{LinGGHD20},
and the box regression loss
(including L1 and GIoU~\cite{RezatofighiTGS019} loss)
is the same as DETR~\cite{CarionMSUKZ20}.

\section{Experiments}

\subsection{Setting}
\noindent\textbf{Dataset.}
We perform the experiments on the COCO $2017$~\cite{LinMBHPRDZ14}
detection dataset.
The dataset contains about $118$K training images and
$5$K validation (\texttt{val}) images.

\begin{table*}[t]
    \centering\setlength{\tabcolsep}{10pt}
        \footnotesize
            \renewcommand{\arraystretch}{1.3}
    \caption{Results for multi-scale and higher-resolution DETR variants.
    We do not expect that our approach performs on par
    as our approach (single-scale, $16 \times$ resolution) does not use a strong multi-scale or $8\times$ resolution encoder.
    Surprisingly,
    the AP scores of our approach with DC$5$-R$50$
    and DC$5$-R$101$
    are close to the two multi-scale and higher-resolution DETR variants.
    }
    \label{tab:comparisontomultiscaledetr}
    \begin{tabular}{lc|cccccccc}
        \shline
        Model & \#epochs & GFLOPs & \#params (M) & AP & AP$_{50}$ & AP$_{75}$ & AP$_{S}$ & AP$_{M}$ & AP$_{L}$ \\
        \shline
        Faster RCNN-FPN-R$50$~\cite{RenHG017} & $36$ & $180$ & $42$ & $40.2$ & $61.0$ & $43.8$ & $24.2$ & $43.5$ & $52.0$  \\ 
        Faster RCNN-FPN-R$50$~\cite{RenHG017} & $108$ & $180$ & $42$ & $42.0$  & $62.1$ & $45.5$ & $26.6$ & $45.5$ & $53.4$ \\
        Deformable DETR-R$50$~\cite{ZhuSLLWD20} & $50$ & $173$ & $40$ & $43.8$ & $62.6$ & $47.7$ & $26.4$ & $47.1$ & $58.0$  \\ 
        TSP-FCOS-R$50$~\cite{SunCYK20} & $36$ & $189$ & $-$ & $43.1$ & $62.3$ & $47.0$ & $26.6$ & $46.8$ & $55.9$ \\
        TSP-RCNN-R$50$~\cite{SunCYK20} & $36$ & $188$ & $-$ & $43.8$ & $63.3$ & $48.3$ & $28.6$ & $46.9$ & $55.7$ \\
        TSP-RCNN-R$50$~\cite{SunCYK20} & $96$ & $188$ & $-$ & $45.0$ & $64.5$ & $49.6$ & $29.7$ & $47.7$ & $58.0$ \\
        \hline 
        Conditional DETR-DC5-R50  & $50$ & $195$ & $44$ &  $43.8$ & $64.4$ &  $46.7$ &  $24.0$ & $47.6$ &  $60.7$  \\
        Conditional DETR-DC5-R50   & $108$ & $195$ & $44$ & $45.1$   & $65.4$   & $48.5$   & $25.3$   & $49.0$   & $62.2$   \\
       
        \hline
        Faster RCNN-FPN-R$101$~\cite{RenHG017} & $36$ & $246$ & $60$ & $42.0$ & $62.5$ & $45.9$ & $25.2$ & $45.6$ & $54.6$ \\
        Faster RCNN-FPN-R$101$~\cite{RenHG017} & $108$ & $246$ & $60$ & $44.0$ & $63.9$ & $47.8$ & $27.2$ & $48.1$ & $56.0$ \\
        TSP-FCOS-R$101$~\cite{SunCYK20} & $36$ & $255$ & $-$ & $44.4$ & $63.8$ & $48.2$ & $27.7$ & $48.6$ & $57.3$ \\
        TSP-RCNN-R$101$~\cite{SunCYK20} & $36$ & $254$ & $-$ & $44.8$ & $63.8$ & $49.2$ & $29.0$ & $47.9$ & $57.1$ \\
        TSP-RCNN-R$101$~\cite{SunCYK20} & $96$ & $254$ & $-$ & $46.5$ & $66.0$ & $51.2$ & $29.9$ & $49.7$ & $59.2$ \\
        \hline
        Conditional DETR-DC5-R101 & $50$ & $262$ & $63$ & $45.0$ &  $65.5$ & $48.4$ & $26.1$ & $48.9$ & $62.8$  \\
        Conditional DETR-DC5-R101 & $108$ & $262$ & $63$ & $45.9$ & $66.8$ & $49.5$ & $27.2$ & $50.3$ & $63.3$ \\
        \shline
    \end{tabular}
    \vspace{-.3cm}
\end{table*}

\vspace{.1cm}
\noindent\textbf{Training.}
We follow the DETR training protocol~\cite{CarionMSUKZ20}.
The backbone is the ImageNet-pretrained model
from TORCHVISION with batchnorm layers fixed,
and the transformer parameters are initialized
using the Xavier initialization scheme~\cite{GlorotB10}.
The weight decay is set to be $10^{-4}$.
The AdamW~\cite{LoshchilovH17} optimizer is used.
The learning rates for the backbone and the transformer
are initially set to be
$10^{-5}$ and $10^{-4}$, respectively. 
The dropout rate in transformer is $0.1$.
The learning rate is dropped by a factor of $10$
after $40$ epochs for $50$ training epochs,
after $60$ epochs for $75$ training epochs,
and
after $80$ epochs for $108$ training epochs.

We use the augmentation scheme same as DETR~\cite{CarionMSUKZ20}:
resize the input image such that
the short side is at least $480$
and at most $800$ pixels
and the long size is at most $1333$ pixels; 
randomly crop the image such that
a training image is cropped with probability $0.5$
to a random rectangular patch. 

\vspace{.1cm}
\noindent\textbf{Evaluation.}
We use the standard COCO evaluation.
We report the average precision (AP),
and the AP scores at $0.50$, $0.75$
and for the small, medium, and large objects. 

\subsection{Results}

\noindent\textbf{Comparison to DETR.}
We compare the proposed conditional DETR 
to the original DETR~\cite{CarionMSUKZ20}.
We follow~\cite{CarionMSUKZ20}
and report the results over
four backbones:
ResNet-$50$~\cite{HeZRS16},
ResNet-$101$,
and their $16\times$-resolution extensions
DC$5$-ResNet-$50$
and
DC$5$-ResNet-$101$.

The corresponding DETR models
are named as
DETR-R$50$,
DETR-R$101$,
DETR-DC5-R$50$,
and DETR-DC5-R$101$,
respectively.
Our models
are named as 
conditional DETR-R$50$,
conditional DETR-R$101$,
conditional DETR-DC5-R$50$,
and conditional DETR-DC5-R$101$,
respectively.

Table~\ref{tab:comparisontodetr} presents
the results from DETR and conditional DETR.
DETR with $50$ training epochs performs much worse than $500$ training epochs.
    Conditional DETR with $50$ training epochs 
    for R$50$ 
    and R$101$ 
    as the backbones performs slightly worse than DETR 
    with $500$ training epochs.
    Conditional DETR with $50$ training epochs
    for DC$5$-R$50$ 
    and DC$5$-R$101$ 
    performs similarly as DETR 
    with $500$ training epochs.
    Conditional DETR for the four backbones with $75/108$ training epochs
    performs better than DETR 
    with $500$ training epochs.
    In summary,
    conditional DETR for high-resolution backbones DC$5$-R$50$
    and DC$5$-R$101$ is $10\times$ faster than the original DETR,
    and for low-resolution backbones 
    R$50$ and R$101$
    $6.67\times$ faster.
    In other words, conditional DETR
    performs better for stronger backbones with better performance.

In addition,
we report the results of 
single-scale DETR extensions:
deformable DETR-SS~\cite{ZhuSLLWD20} 
and UP-DETR~\cite{DaiCLC20} in Table~\ref{tab:comparisontodetr}.
Our results over R$50$ and DC$5$-R$50$ are
better than deformable DETR-SS:
$40.9$ vs. $39.4$
and $43.8$ vs. $41.5$.
The comparison might not be fully fair as
for example parameter and computation complexities
are different,
but it implies that
the conditional cross-attention mechanism is beneficial.
Compared to UP-DETR-R$50$,
our results with fewer training epochs 
are obviously better.

\vspace{0.1cm}
\noindent\textbf{Comparison to multi-scale and higher-resolution DETR variants.}
We focus on accelerating the DETR training,
without addressing the issue of high computational complexity in the encoder.
We do not expect that our approach achieves 
on par with DETR variants w/ multi-scale attention and $8\times$-resolution encoders,
e.g.,
TSP-FCOS and TSP-RCNN~\cite{SunCYK20} and deformable DETR~\cite{ZhuSLLWD20},
which are able to reduce the encoder computational complexity
and improve the performance due to multi-scale and higher-resolution. 

The comparisons in Table~\ref{tab:comparisontomultiscaledetr} surprisingly show that our approach on DC$5$-R$50$ ($16 \times$) performs 
same as deformable DETR-R$50$ (multi-scale, $8\times$).
Considering that the AP of the single-scale deformable DETR-DC$5$-R$50$-SS is $41.5$ (lower than ours $43.8$) (Table~\ref{tab:comparisontodetr}),
one can see that deformable DETR benefits a lot from  
the multi-scale and higher-resolution encoder
that potentially benefit our approach,
which is currently not our focus 
and left as our future work.

The performance of our approach 
is also on par with TSP-FCOS and TSP-RCNN.
The two methods contain 
a transformer encoder
over a small number of selected positions/regions
(feature of interest in TSP-FCOS
and region proposals in TSP-RCNN)
without using the transformer decoder,
are extensions of FCOS~\cite{TianSCH19} and Faster RCNN~\cite{RenHG017}.
It should be noted that position/region selection
removes unnecessary computation in self-attention
and reduces computation complexity dramatically.

\subsection{Ablations}
\noindent\textbf{Reference points.}
We compare three ways 
of forming reference points $\mathbf{s}$:
(i) $\mathbf{s} = (0, 0)$, same to the original DETR,
(ii) learn $\mathbf{s}$ as model parameters
and each prediction is associated with different reference points,
and (iii) predict each reference point 
$\mathbf{s}$ from the corresponding object query.
We conducted the experiments 
with ResNet-$50$ as the backbone.
The AP scores are $36.8$, $40.7$,
and $40.9$,
suggesting that (ii) and (iii) perform on par
and better than (i).

\begin{table}[t]
    \centering
    \setlength{\tabcolsep}{8.4pt}
        \footnotesize
            \renewcommand{\arraystretch}{1.3}
    \caption{\small Ablation study for the ways forming the conditional spatial query.
    CSQ = our proposed conditional spatial query scheme.
    Please see the first two paragraphs in Section 5.3
    for the meanings of CSQ variants.
    Our proposed CSQ manner performs better.
    The backbone ResNet-$50$ is adopted.}
    \label{tab:ablationstudy}
    \begin{tabular}{ c | c | c | c  |  c | c}
    \shline
        Exp.  &  CSQ-C & CSQ-T & CSQ-P & CSQ-I & CSQ \\
        \shline
        GFLOPs &  $89.3$     &   $89.5$    &   $89.3$  & $89.5$  &  $89.5$   \\
        \hline
        AP &   $37.1$    &  $37.6$     &   $37.8$ & $ 40.2 $  & $40.9$    \\
        \shline
    \end{tabular}
    \vspace{-.3cm}
\end{table}

\vspace{.1cm}
\noindent\textbf{The effect of the way
forming the conditional spatial query.}
We empirically study how 
the transformation $\boldsymbol{\uplambda}_q$ 
and 
the positional embedding $\mathbf{p}_s$
of the reference point,
used to form the conditional spatial query
$\mathbf{p}_q = \boldsymbol{\uplambda}_q \odot \mathbf{p}_s$,
make contributions to
the detection performance.

We report the results of our conditional DETR,
and 
the other ways forming the spatial query
with:
(i) CSQ-P - only the positional embedding $ \mathbf{p}_s$, 
(ii) CSQ-T - only the transformation $\boldsymbol{\uplambda}_q$,
(iii) CSQ-C - the decoder content embedding $ \mathbf{f}$,
and (iv) CSQ-I - the element-wise product of the transformation predicted from the decoder self-attention output $\mathbf{c}_q$ and the positional embedding $\mathbf{p}_s$.
The studies in Table~\ref{tab:ablationstudy} imply that
our proposed way (CSQ) 
performs overall the best,
validating our analysis about the transformation predicted from the decoder embedding and the positional embedding of the reference point
in Section~\ref{sec:conditionalcrossattention}.

\vspace{.1cm}
\noindent\textbf{Focal loss and offset regression
with respect to learned reference point.}
Our approach follows deformable DETR~\cite{ZhuSLLWD20}:
use the focal loss with $300$ object queries to form
the classification loss
and predict the box center
by regressing the offset with respect to 
the reference point.
We report how the two schemes affect the DETR performance 
in Table~\ref{tab:lossfunctions}.
One can see that separately using the focal loss
or center offset regression
without learning referecence points
leads to a slight AP gain
and combining them together leads to a larger AP gain.
Conditional cross-attention in our approach built on the basis of focal loss and offset regression
brings a major gain $4.0$.

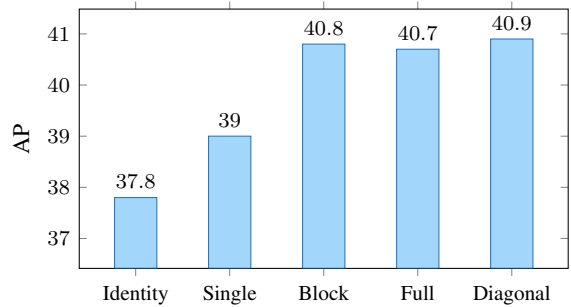
\begin{figure}[t]
\begin{tikzpicture}[baseline]
\pgfplotsset{set layers, compat=1.3, 
every axis/.append style={
font=\footnotesize,
}
}
\begin{axis}[
	footnotesize,
	scale only axis,
	ybar,
	enlargelimits=0.15,
	symbolic x coords={
	Identity, Single, Block, Full, Diagonal},
	x post scale=1.3,
	y post scale=.8,
    ytick distance=1,
    ymin=37,
	xtick=data,
    nodes near coords, 
	nodes near coords align={vertical},
    ylabel={AP},
    bar width=16pt,
]
\addplot[fill=mayablue, draw=mediumelectricblue, fill opacity=0.66, draw opacity=0.8, text opacity=1]
	coordinates{
	(Identity, 37.8)
	(Single, 39.0)
	(Block, 40.8)
	(Full, 40.7)
	(Diagonal, 40.9)
	};
\end{axis}
\end{tikzpicture}
\caption{The
empirical results
for different forms of linear projections that 
are used to compute the spatial queries
for conditional multi-head cross-attention.
Diagonal (ours), Full, and Block 
perform on par.
The backbone ResNet-$50$ is adopted.
}
\label{fig:conditionalspatialqueries}
\vspace{-.46cm}
\end{figure}

\vspace{.1cm}
\noindent\textbf{The effect of linear projections $\mathbf{T}$
forming the transformation.}
Predicting the conditional spatial query
needs to learn the linear projection $\mathbf{T}$
from the decoder embedding
(see Equation~\ref{eqn:conditionalspatialembedding}). 
We empirically study how 
the linear projection forms
affect the performance.
The linear projection forms include:
an \emph{identity} matrix that means not to learn the linear projection,
a \emph{single} scalar,
a \emph{block} diagonal matrix
meaning that each head has a learned $32\times 32$ linear projection matrix,
a \emph{full} matrix without constraints,
and a \emph{diagonal} matrix.
Figure~\ref{fig:conditionalspatialqueries}
presents the results.
It is interesting
that a single-scalar helps improve the performance,
maybe due to narrowing down the spatial range to the object area.
Other three forms, \emph{block} diagonal, \emph{full},
and \emph{diagonal} (ours), perform on par.

\begin{table}[t]
    \centering
    \setlength{\tabcolsep}{15.5pt}
        \footnotesize
            \renewcommand{\arraystretch}{1.3}
    \caption{\small The empirical results 
    about the focal loss (FL), offset regression (OR) for box center prediction, and our conditional spatial query (CSQ).
    The backbone ResNet-50 is adopted.}
    \label{tab:lossfunctions}
    \begin{tabular}{ c  c  c | c | c}
    \shline
         OR & FL & CSQ & GFLOPs & AP \\
        \shline
          &  &  & $85.5$ & $34.9$ \\
        \cmark &  &  & $85.5$ & $35.0$ \\
         & \cmark &  & $88.1$ & $35.3$ \\
        \cmark & \cmark &  & $88.1$ & $36.9$ \\
        \cmark & \cmark & \cmark & $89.5$ & $40.9$ \\
        \shline
    \end{tabular}
    
\end{table}

\section{Conclusion}
We present a simple conditional cross-attention mechanism.
The key is to learn a spatial query from the corresponding reference point and decoder embedding.
The spatial query
contains the spatial information
mined for the class and box prediction 
in the previous decoder layer,
and 
leads to spatial attention weight maps
highlighting the bands containing extremities
and small regions inside the object box.
This shrinks the spatial range for the content query
to localize the distinct regions,
thus relaxing the dependence on the content query and reducing the training difficulty.
In the future, we will study the proposed conditional cross-attention mechanism
for human pose estimation~\cite{GengSXZW21,Wang2019,SunXLW19} and line segment detection~\cite{XuXCT21}.

\vspace{.1cm}
\noindent \textbf{Acknowledgments.} We thank the anonymous reviewers for their insightful comments and suggestions on our manuscript.

{\small
 \bibliographystyle{ieee_fullname}
 \bibliography{detection}
 }
 
\end{document}